\documentclass[runningheads]{llncs}
\usepackage{graphicx}
\usepackage{float}
\usepackage[acronym,toc]{glossaries} 
\newacronym{fl}{FL}{Federated Learning}
\newacronym{ml}{ML}{Machine Learning}
\newacronym{ai}{AI}{Artificial Intelligence}
\newacronym{api}{API}{Application Programming Interface}
\newacronym{fc}{FC}{Federated Core}
\newacronym{tff}{TFF}{TensorFlow Federated}
\newacronym{qoi}{QoI}{Quality of Information}
\newacronym{smc}{SMC}{Secure Multi-party Computation}
\newacronym{dp}{DP}{Differential Privacy}
\newacronym{he}{HE}{Homomorphic Encryption}
\newacronym{iot}{IoT}{Internet of Things}
\newacronym{hvac}{HVAC}{Heating Ventilation Air Condition}
\newacronym{ifl}{IFL}{Industrial Federated Learning}
\newacronym{qos}{QoS}{Quality of Service}
\newacronym{sdk}{SDK}{Software Development Kit}

\begin{document}
\title{Industrial Federated Learning --\\ Requirements and System Design}
%
%

\author{Thomas Hiessl\inst{1,2}\orcidID{0000-0003-0293-0159} \and
Daniel Schall\inst{1}\orcidID{0000-0002-1038-0241} \and \\
Jana Kemnitz\inst{1}\orcidID{0000-0003-0342-4952} \and
Stefan Schulte\inst{2}\orcidID{0000-0001-6828-9945}}
\authorrunning{T. Hiessl et al.}
%
\institute{Siemens Corporate Technology, 1210 Vienna, Austria \\
\email{\{hiessl.thomas, daniel.schall, jana.kemnitz\}@siemens.com} \and
Vienna University of Technology, Karlsplatz 13, 1040 Vienna, Austria 
\email{s.schulte@dsg.tuwien.ac.at}}
\maketitle             
\begin{abstract}
\acrfull{fl} is a very promising approach for improving decentralized \acrfull{ml} models 
by exchanging knowledge between participating clients without revealing private data. 
Nevertheless, \acrshort{fl} is still not tailored to the industrial context as strong data similarity is assumed for all \acrshort{fl} tasks.
This is rarely the case in industrial machine data with variations 
in machine type, operational- and environmental conditions. 
Therefore, we introduce an \acrfull{ifl} system supporting knowledge exchange in continuously evaluated and 
updated FL cohorts of learning tasks with sufficient data similarity. 
This enables optimal collaboration of business partners in common \acrshort{ml} problems, 
prevents negative knowledge transfer, 
and ensures resource optimization of involved edge devices. 

\keywords{Federated Learning  \and Industrial AI \and Edge Computing.}
\end{abstract}

%
\section{Introduction}
\label{sec:introduction}  

Industrial manufacturing systems often consist of various operating machines and automation systems. 
High availability and fast reconfiguration of each operating machine is key to 
frictionless production resulting in competitive product pricing ~\cite{parunakDAI1996}. 
To ensure high availability of each machine, 
often condition monitoring is realized based on \acrfull{ml} models deployed to edge devices, e.g., indicating anomalies in production ~\cite{conditionMonitoringHusakovic2018}. 
The performance of these \acrshort{ml} models clearly depends on available training data, 
which is often only available to a limited degree for individual machines. 
Increasing training data might be realized by sharing data within the company or with an external industry partner~\cite{dataIntegrationPrivacyPreserving}.
The latter approach is often critical as vulnerable business or privact information might be contained.

The recently emerged Federated Learning \acrshort{fl} method enables to train a \acrshort{ml} model on multiple 
local datasets contained in local edge devices without exchanging data samples~\cite{privacyPreservingDLShokri}. 
In this privacy-preserving approach, typically a server receives parameters (e.g., gradients or weights of neural networks) 
from local models trained on decentralized edge devices 
and averages these parameters to build a global model~\cite{mcmahan2016communicationefficient}.
After that, the averaged global model parameters are forwarded to edge devices to update local models.
This process is repeatedly executed until the global model converges or a defined break-up condition is met. 

\begin{figure}[tb]
    \includegraphics[width=\textwidth]{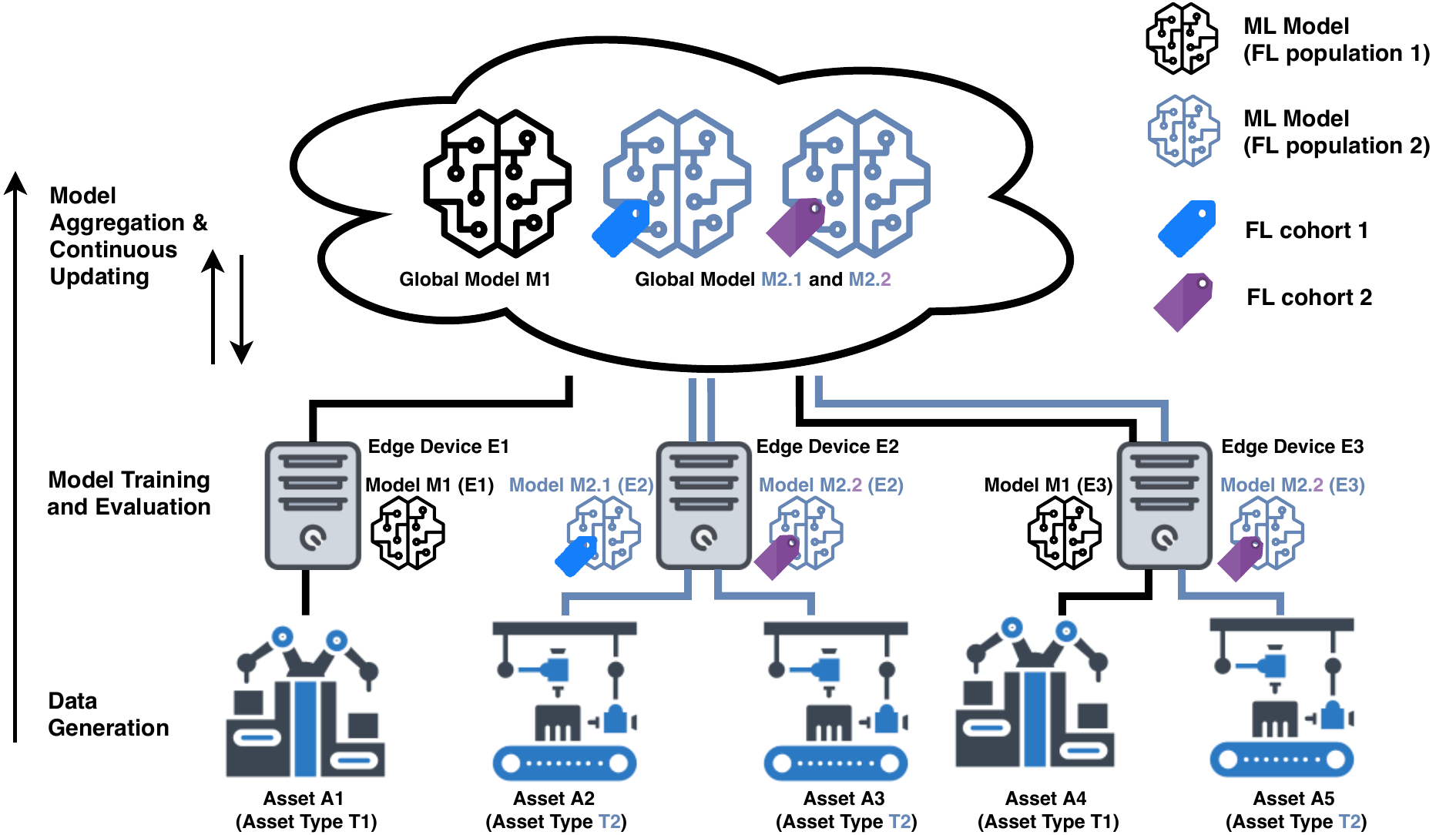}
    \caption{\acrfull{fl} with industrial assets; 
    Assets generate data that are used in learning tasks for \acrshort{ml} models executed on edge devices;
    Learning tasks for \acrshort{ml} models based on the same asset type are part of a \acrshort{fl} population;
    Learning tasks for \acrshort{ml} models with similar data are part of a \acrshort{fl} population subset named \acrshort{fl} cohort;
    Knowledge transfer in continuously evaluated and updated \acrshort{fl} cohorts ensures optimal collaboration with respect to model performance and business partner criteria} \label{fig:scenario}
\end{figure}

However, to solve the discussed challenges of successfully applying \acrshort{ml} models in industrial domains, 
\acrshort{fl} needs to be adapted.
Therefore, the integration of operating machines and its digital representations named  
\emph{assets}\footnote{https://documentation.mindsphere.io/resources/html/asset-manager/en-US/113537583883.html}
need to be considered as depicted in Figure~\ref{fig:scenario}. 
Assets generate data on the shop floor during operation.
Edge devices record this data to enable training of \acrshort{ml} models 
e.g., in the field of anomaly detection aiming to identify abnormal behavior of machines in production.
To improve the model quality, \acrshort{fl} is applied by aggregating model parameters centrally in a global model, e.g., in the cloud, 
and sending out updates to other edge devices. 
Typically, all models of local learning tasks corresponding to the same \acrshort{ml} problem are updated. 
This set of tasks is called a \acrshort{fl} population. 
In the depicted industry scenario, 
a \acrshort{fl} population corresponds to all learning tasks for models trained on asset data with same data scheme, 
which is typically ensured if assets are of the same asset type, 
e.g., learning tasks of models \emph{M2.1~(E2)}, \emph{M2.2~(E2)}, and \emph{M2.2~(E3)} belong to \emph{FL~population~2}, since they are based on assets of \emph{Asset Type T2}. 
In contrast, learning tasks of models \emph{M1~(E1)} and \emph{M1~(E3)} belong to \emph{FL population 1}.
However, assets even of same asset type could face heterogenous environmental and operation conditions which affect recorded data.
Due to these potential dissimilarities in asset data, 
negative knowledge transfer can be caused by the model updates which decreases model performance~\cite{torrey2010transfer}.
For this, industrial \acrshort{fl} systems need to consider \acrshort{fl} cohorts as subsets of a \acrshort{fl} population.
This enables knowledge sharing only within e.g., \emph{FL~cohort~2} including \emph{M2.2} models using similar asset data. 

For this, we propose to establish \acrshort{fl} system support 
for knowledge exchange in \acrshort{fl} cohorts involving \acrshort{ml} models based on asset data from industry. 
Furthermore, it needs support for continuous adaption of \acrshort{fl} cohorts as \acrshort{ml} models evolve over time.
To additionally support efficient \acrshort{fl} with high quality of asset data, 
we aim for resource optimization of involved edge devices and appropriate consideration of \acrfull{qoi} metrics~\cite{AIMQ-InfoQuality-Lee-2002}. 
Hence, our contribution comprises requirements and a system design for \acrfull{ifl} which we introduce in this paper.
\acrshort{ifl} aims to improve collaboration on training and evaluating \acrshort{ml} models in industrial environments. 
For this, we consider current \acrshort{fl} systems and approaches~\cite{FLSystemDesign-Bonawitz-2019,AsynchronousFLforEdge-Chen-2019,LFRL-Liu-2019,mcmahan2016communicationefficient,privacyPreservingDLShokri} 
and incorporate industry concepts as well as experience from industrial projects.
The design of the \acrshort{ifl} system is presented with respect to supported workflows, domain model, and architecture. 

In Section~\ref{sec:flNotation} we refer to the basic notation of \acrshort{fl}.
We review related work in Section~\ref{sec:relatedWork} 
and subsequently present requirements of \acrshort{ifl} in Section~\ref{sec:requirements}.
The design of the~\acrshort{ifl} system is presented in Section~\ref{sec:systemDesign} 
with respect to supported workflows, domain models, and architectures.
We conclude in Section~\ref{sec:conclusion} and provide an outline to future work

\section{IFL Notation}
\label{sec:flNotation}
To introduce the basic notation of an \acrshort{ifl} systems, we extend the \acrshort{fl} notation by 
Bonawitz~et~al.~\cite{FLSystemDesign-Bonawitz-2019} that define \emph{device}, \emph{FL server}, \emph{FL task}, \emph{FL population} and \emph{FL plan}.
Devices are hardware platforms as e.g., industrial edge devices or mobile phones, 
running \emph{FL clients} to execute the computation necessary for training and evaluating \acrshort{ml} models.
To use \acrshort{fl}, a \acrshort{fl} client communicates to the \acrshort{fl} server to run \acrshort{fl} tasks for a given \acrshort{fl} population.
The latter one is a globally unique name that identifies a learning problem which multiple \acrshort{fl} tasks have in common.
The \acrshort{fl} server aggregates results (i.e., model updates), persists the global model, and provides it to \acrshort{fl} clients of a given \acrshort{fl} population. 
A \acrshort{fl} plan corresponds to a FL task and represents its federated execution instructions for the \acrshort{fl} server and involved \acrshort{fl} clients. 
It consists of sequences of \acrshort{ml} steps as e.g., data pre-processing, training, and evaluation to be executed by \acrshort{fl} clients 
and instructions for aggregating \acrshort{ml} models on the \acrshort{fl} server. 
Furthermore, we define \emph{FL cohorts} that group multiple \acrshort{fl} tasks within the same \acrshort{fl} population 
and with similarities in their underlying asset data.

\section{Related Work}
\label{sec:relatedWork}

\subsection{FL Systems}
Most of the current \acrshort{fl} studies focus on federated algorithm design and efficiency improvement~\cite{surveyFL-2019}.
Besides that, Bonawitz et al.~\cite{FLSystemDesign-Bonawitz-2019} built a scalable production system for \acrshort{fl} 
aiming to facilitate learning tasks on mobile devices using \emph{TensorFlow}\footnote{https://www.tensorflow.org/}.
Furthermore, \emph{NVIDIA Clara}\footnote{https://devblogs.nvidia.com/federated-learning-clara/} provided an SDK 
to integrate custom \acrshort{ml} models in a \acrshort{fl} environment. 
This system has been evaluated with data from the medical domain, considering a scenario with decentralized image datasets located in hospitals.
However, no aspects of dynamically changing data patterns in learning tasks of \acrshort{fl} cohorts have been considered in literature so far. 

\subsection{Client Selection}
\label{subsec:clientSelection}
Nishio~et~al.~\cite{ClientSelectionFL-Nishio-2019} optimize model training duration in \acrshort{fl} by selecting only a subset of \acrshort{fl} clients. 
Since they face heterogeneous conditions and are provisioned with diverse resource capabilities, 
not all \acrshort{fl} clients will manage to deliver results in decent time. 
For this, only those who deliver before a deadline are selected in the current training round.
To achieve the best accuracy for the global model, 
the \acrshort{fl} server may select \acrshort{fl} clients based on their model evaluation results on held out validation data~\cite{FLSystemDesign-Bonawitz-2019}.
This allows to optimize the configuration of \acrshort{fl} tasks 
such as centrally setting hyperparameters for model training or defining optimal number of involved \acrshort{fl} clients.
Although, in \acrshort{ifl} these client selection approaches need to be considered, 
the \acrshort{ifl} system further selects \acrshort{fl} clients based on collaboration criteria 
with respect to potential \acrshort{fl} business partners. 

\subsection{Continuous Federated Learning}
Liu~et~al.~\cite{LFRL-Liu-2019} propose a cloud-based FL system for reinforcement tasks of robots navigating around obstacles. 
Since there exist robots that train much and therefore update \acrshort{ml} models continuously, 
the authors identify the need for sharing these updates with other federated robots. 
These updates are asynchronously incorporated in the global model to eventually enhance navigation skills of all involved robots.
Based on that, in \acrshort{ifl} the continuous updates are used to re-evaluate 
data similarity that is needed to ensure high model quality within a \acrshort{fl} cohort organization.

\section{Requirements}
\label{sec:requirements}
In this section we now present requirements that should be coverd by an \acrshort{ifl} system. 
Based on \acrshort{fl} system features discussed in~\cite{surveyFL-2019}, 
we add requirements with respect to industrial data processing and continuous adaptation of the system.

\subsection{Industrial Metadata Management}
\label{subsec:industrialMetadata}
To support collaboration of \acrshort{fl} clients, we identify the requirement of publishing metadata describing the organization and its devices. 
Based on this, \acrshort{fl} clients can provide criteria for collaborating with other selected \acrshort{fl} clients. 
Although actual raw data is not shared in \acrshort{fl}, it enables to adhere to company policies for interacting with potential partners.
Asset models as provided by Siemens MindSphere\footnote{https://documentation.mindsphere.io/resources/html/asset-manager/en-US/113537583883.html}
describes the data scheme for industrial \acrfull{iot} data. 
Since industrial \acrshort{fl} clients target to improve machine learning models using asset data, 
metadata describing the assets builds the basis for collaborating in suitable \acrshort{fl} populations.

\subsection{FL Cohorts}
As discussed in Section~\ref{subsec:clientSelection}, \acrshort{fl} client selection plays 
a role in \acrshort{fl} to reduce duration of e.g., training or evaluation~\cite{ClientSelectionFL-Nishio-2019}.
Furthermore, client selection based on evaluation using held-out validation data, can improve accuracy of the global model~\cite{FLSystemDesign-Bonawitz-2019}. 
In our experience, these approaches do not sufficiently address data generated by industrial assets and processed by ~\acrshort{fl} clients. 
For this, our approach aims for considering asset data characteristics 
for achieving optimal accuracy and performance for all individual client models. 
To this end, we identify the requirement of evaluating models in regards to similarities of asset data influenced 
by operating and environmental conditions.
This is the basis for building \acrshort{fl} cohorts of \acrshort{fl} tasks using asset data with similar characteristics.
\acrshort{fl} cohorts enable that \acrshort{fl} clients only share updates within a subset of \acrshort{fl} clients,
whose submitted \acrshort{fl} tasks belong to the same \acrshort{fl} cohort. 
These updates probably improve their individual model accuracy better, 
as if updates would be shared between \acrshort{fl} clients that face very heterogeneous data 
due to e.g., different environmental or operating conditions of involved assets. 
In manufacturing industries there are situations where assets are placed in sites with similar conditions,  
as, e.g., placing production machines into shop floors with similar temperature, noise and other features considered in the model prediction.
In such cases, the \acrshort{ifl} system needs to build \acrshort{fl} cohorts. 

\subsection{Quality of Information}
\label{subsec:qoi}
Since each \acrshort{fl} client trains and evaluates on its local data set, 
aggregated global models result from data sets with diverse \acrshort{qoi}. 
Furthermore, due to different agents operating in the industry
as e.g., fully autonomous control systems as well as semi-autonomous ones with human interaction~\cite{archon1994}, 
different data recording approaches can influence \acrshort{qoi} of asset data sets. 
Lee et al.~\cite{AIMQ-InfoQuality-Lee-2002} discuss different dimensions of \acrshort{qoi} 
as e.g., \emph{free-of-error}, \emph{relevancy}, \emph{reputation}, \emph{appropriate amount}, \emph{believability}, 
\emph{consistent representation} and \emph{security}. 
Based on that, we derive that there is the need to evaluate \acrshort{qoi} on \acrshort{fl} clients 
and use resulting metrics on the \acrshort{fl} server to decide on the extent of contribution 
of an individual \acrshort{fl} client in the parameter aggregation process.
Storing \acrshort{qoi} metrics next to existing industrial metadata of participating organizations 
further enhances building and updating suitable \acrshort{fl} cohorts.

\subsection{Continuous Learning}
\acrfull{ai} increasingly enables operation of industrial processes to realize flexibility, efficiency, and sustainability. 
For this, often domain experts have to repeatedly understand new data with respect to its physical behavior 
and the meaning of parameters of the underlying process~\cite{lee2018IndustrialAI}.
Moreover, continuously involving domain experts and data scientists in updating \acrshort{ml} models 
by e.g., providing labels to recently recorded time series data, 
is a resource-intensive process, that can be faciliated by continuously collaborating in \acrshort{fl}. 
Based on that, we identify the need of supporting continuously re-starting FL learning processes and cohort reorganization over time 
to consider major changes in asset time series data.

%

\subsection{Scheduling and Optimization}
Executing \acrshort{fl} plans can cause heavy loads on edge devices, as 
e.g., training of \acrshort{ml} models on large data sets~\cite{FLSystemDesign-Bonawitz-2019}. 
Bonawitz et al.~\cite{FLSystemDesign-Bonawitz-2019} identified the need for device scheduling. 
This involves that, e.g., multiple \acrshort{fl} plans are not executed in parallel on single devices with little capacities, 
or that repeated training on older data sets is avoided while training on \acrshort{fl} clients with new data is promoted.   
For industry purposes, it further needs optimization of cohorts communication. 
This means, that \acrshort{fl} tasks linked to a \acrshort{fl} cohort, 
can be transferred to other cohorts if this improves communication between 
involved \acrshort{fl} clients with respect to e.g., latency minimization~\cite{hiesslDSPICFEC2019}. 
We believe, this decreases model quality due to preferring communication metrics over model quality metrics. 
However, \acrshort{ifl} systems need to consider this trade-off in an optimization problem and solve it to maximize overall utility.
Furthermore, collaboration restrictions of \acrshort{fl} clients needs to be considered in the optimization problem. 
This ensures that no organization joins \acrshort{fl} cohorts with other organizations that they do not want to collaborate with.

\section{System Design}
\label{sec:systemDesign}

\begin{figure}[tb]
    \includegraphics[width=\textwidth]{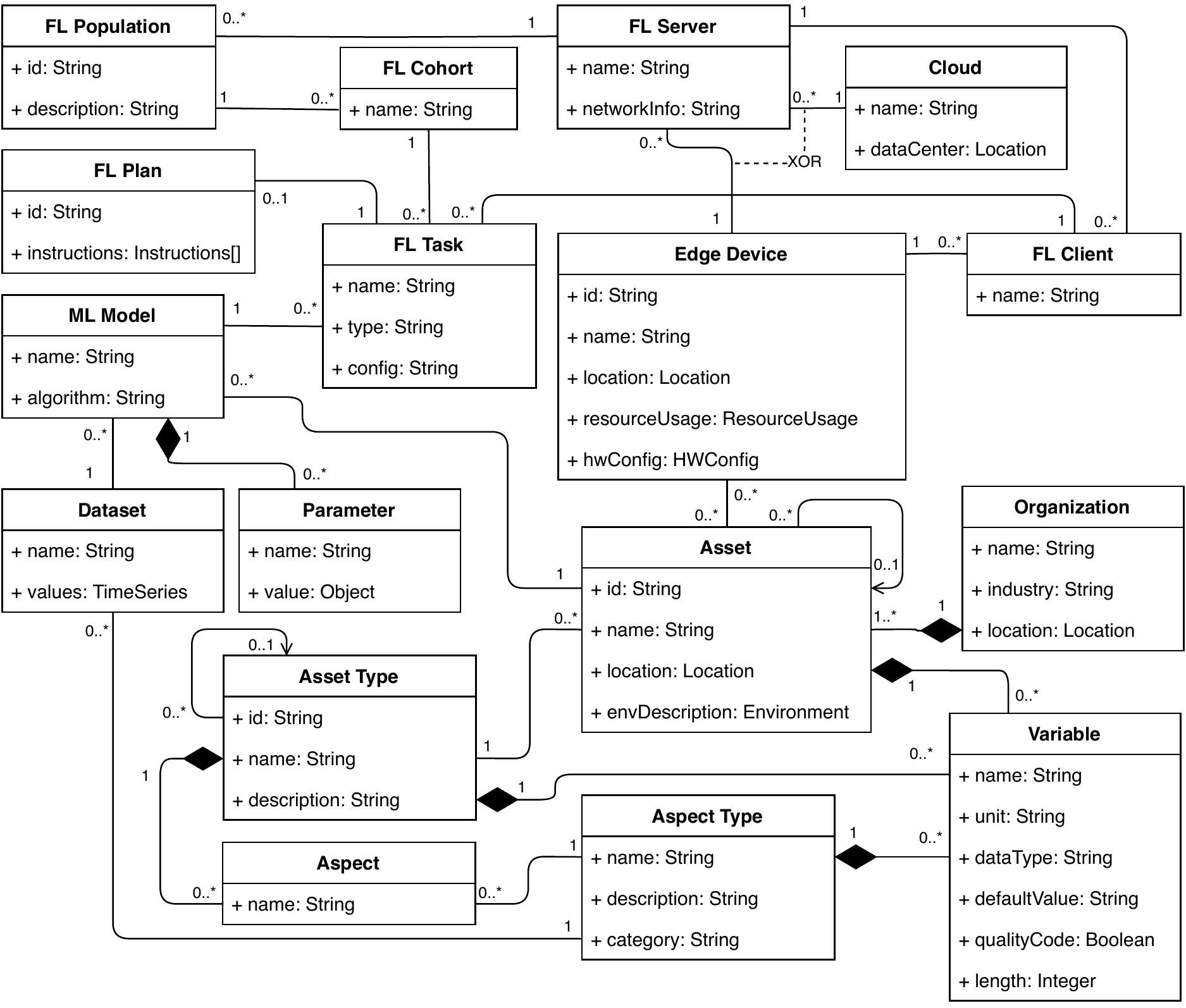}
    \caption{Domain Model} \label{fig:domainModel}
\end{figure}

\subsection{Domain Model}
\label{subsec:domainModel}
To establish a domain model for \acrshort{ifl}, we consider FL terminology~\cite{FLSystemDesign-Bonawitz-2019} 
as well as concepts from industrial asset models as discussed in Section~\ref{subsec:industrialMetadata}.
For this, Figure~\ref{fig:domainModel} depicts \emph{FL Population}, \emph{FL Server}, \emph{FL Client}, \emph{FL Task} and \emph{FL Plan} 
as discussed in Section~\ref{sec:flNotation}. 
Herein, we consider to deploy and run the \acrshort{fl} server either in the \emph{Cloud} or on an \emph{Edge Device}.
The \acrshort{fl} client is hosted on an industrial edge device, that is a hardware device on a given \emph{location}.
To support scheduling and optimization decisions of the \acrshort{fl} server, 
the edge device contains \emph{resource usage} metrics and hardware specifications (\emph{hwConfig}).
A \acrshort{fl} task refers to a \emph{ML Model} that needs to be trained with an \emph{algorithm} 
on a given \emph{Dataset} consisting of time series~\emph{values}. 
The scheme of the Dataset is defined by an \emph{Aspect Type}, which contains a set of \emph{Variables}.
Each Variable has \emph{name}, \emph{unit}, \emph{dataType}, \emph{defaultValue} and \emph{length} attributes
to define the content of the corresponding time series values.
The \emph{qualityCode} indicates wheter a variable supports \emph{OPC Quality Codes}\footnote{https://www.opcsupport.com/s/article/What-are-the-OPC-Quality-Codes}.
This enables to record and evaluate \acrshort{qoi} metrics on the \acrshort{fl} client as discussed in Section~\ref{subsec:qoi}.
Since industrial \acrshort{ml} tasks typically consider data from industrial assets, 
we define an \emph{Asset} (e.g., a concrete engine) operating on a given \emph{location} facing environmental conditions (\emph{envDescription}).
The asset is an instance of an \emph{Asset Type} (e.g., an engine) that collects multiple \emph{aspects} (e.g., surface vibrations) 
of corresponding aspect types (e.g., vibration) again collecting variables (e.g., vibrations in x,y,z dimensions). 
The asset is connected to an edge device which is recording data for it.
To express the complexity of industrial organizations, hierarchical asset structures can be built
as it is depicted with recursive associations of assets and related asset types, 
considering nesting of, e.g., overall shop floors, their assembly lines, involved machines and its parts.
Finally, we introduce \acrshort{fl} cohorts as groups of \acrshort{fl} tasks. 
A \acrshort{fl} cohort is built with respect to similarities of assets considered in the attached \acrshort{ml} model. 
So, creating \acrshort{fl} tasks intents to typically solve ML problems based on asset data, 
whereas the aspect type referred in the Dataset of the ML model are used in the linked asset.

\subsection{Workflows}
To regard the requirements of Section~\ref{sec:requirements}, we propose several workflows to be supported by the \acrshort{ifl} system.

\subsubsection{FL Client Registration}
Assuming the \acrshort{fl} server to be in place, the \acrshort{fl} client starts participation in the \acrshort{ifl} system by registering
itself. For this, the \acrshort{fl} client has to submit a request including organization and edge device information. 
Furthermore, aspect types are handed in, describing the data scheme 
based on which the organization is willing to collaborate in \acrshort{fl} processes with other organizations.
Additionally, the assets enabled for \acrshort{fl} are posted to the \acrshort{fl} server, 
to provide an overview to other organizations and to ensure that IFL can build \acrshort{fl} cohorts based on respective environmental conditions. 

\subsubsection{Cohort Search Criteria Posting}
After \acrshort{fl} client registration, other \acrshort{fl} clients can request a catalog of edge devices, 
organizations and connected assets. 
Based on this, cohort search criteria can be created potentially including organizations, 
industries, and asset types as well as aspect types. 
This enables to match submitted \acrshort{fl} tasks to \acrshort{fl} cohorts based on client restrictions for collaboration and their \acrshort{ml} models.

\subsubsection{Submit and Run FL Tasks}
The \acrshort{fl} client creates a \acrshort{fl} task including references to the \acrshort{ml} model without revealing the actual data set
and submits it to the \acrshort{fl} server. If \acrshort{fl} tasks target the same problems, 
i.e., reference to the same aspect types and corresponding \acrshort{ml} model, 
the provided \acrshort{fl} task is attached to an existing \acrshort{fl} population, otherwise a new \acrshort{fl} population is created.
\acrshort{ifl} then builds \acrshort{fl} cohorts of \acrshort{fl} tasks based on metadata provided during registration and posted cohort search criteria.
If no cohort search criteria is provided by the \acrshort{fl} client, 
the submitted \acrshort{fl} Tasks are initially considered in the \emph{default} \acrshort{fl} cohort of the given \acrshort{fl} population. 
To actually start \acrshort{fl}, a \acrshort{fl} plan is created including server and client instructions to realize 
e.g., \emph{Federated Averaging}~\cite{mcmahan2016communicationefficient} on the server 
and training of \acrshort{ml} models on every involved \acrshort{fl} client.
The \emph{configuration} of \acrshort{fl} tasks allows for defining parameters for supported algorithms of \acrshort{ifl}
for, e.g., setting break-up conditions for \acrshort{fl} or defining the number of repeated executions over time.
Since \acrshort{fl} tasks are either realized as training or evaluation plan, the exchanged data between \acrshort{fl} client and \acrshort{fl} Server are different.
While training plans typically include the sharing of model parameters as, e.g., gradients or weights of neural networks, 
evaluation plan execution results in metrics that are stored by \acrshort{ifl} to further enable \acrshort{fl} cohort reconfiguration and optimization.

\subsubsection{Update FL Cohorts}
Collected metrics in the \acrshort{fl} process enable to update \acrshort{fl} cohorts 
with respect to splitting and merging \acrshort{fl} cohorts. 
Furthermore, moving \acrshort{fl} tasks between cohorts is considered in \acrshort{ifl}.
The respective metrics include information like the environmental changes of assets and model accuracy. 
Furthermore, similarity measures of \acrshort{ml} models are computed based on possible server-provided data. 
If such evaluation data is present, a strategy for updating \acrshort{fl} cohorts includes 
to put \acrshort{fl} tasks in the same \acrshort{fl} cohort, where its \acrshort{ml} model predicts ideally the same output based on provided input samples. 

\subsubsection{Evaluate QoI}
The \acrshort{qoi} of raw data used by each \acrshort{fl} client is computed on edge devices 
and mapped to OPC Quality Codes as defined in Section~\ref{subsec:domainModel}. 
Besides using submitted \acrshort{qoi} for e.g., updating \acrshort{fl} cohorts, 
\acrshort{ifl} considers \acrshort{qoi} in the contribution weights of \acrshort{fl} clients 
when it comes to weighted averaging of model parameters as defined in~\cite{mcmahan2016communicationefficient}.

\subsubsection{Continuous Learning}
After time series data is updated and if needed properly labelled, \acrshort{fl} tasks are submitted. 
For this, either synchronous~\cite{mcmahan2016communicationefficient} or asynchronous~\cite{AsynchronousFLforEdge-Chen-2019} \acrshort{fl} processes are triggered.
In the asynchronous case, \acrshort{ifl} determines the timing for notifying \acrshort{fl} clients
to update \acrshort{ml} models according to recent improvements of one \acrshort{fl} client.

\subsubsection{Optimize Computation and Communication}
First, the \acrshort{fl} server loads resource usage from edge devices to determine the load caused by executed processes.
Second, network statistics (e.g., latency) are identified as recorded for model update sharings between \acrshort{fl} clients and the \acrshort{fl} Server. 
Third, statistics of past \acrshort{fl} plan executions, e.g., duration of processing is loaded to be incorported in an optimization model.
Finally, this model optimizes future \acrshort{fl} plan executions 
considering \acrshort{qos} criteria~\cite{hiesslDSPICFEC2019} as processing cost, network latency, 
and cohort reconfiguration cost.

\subsection{Architecture}
\label{subsec:architecture}
\begin{figure}[tb]
    \includegraphics[width=\textwidth]{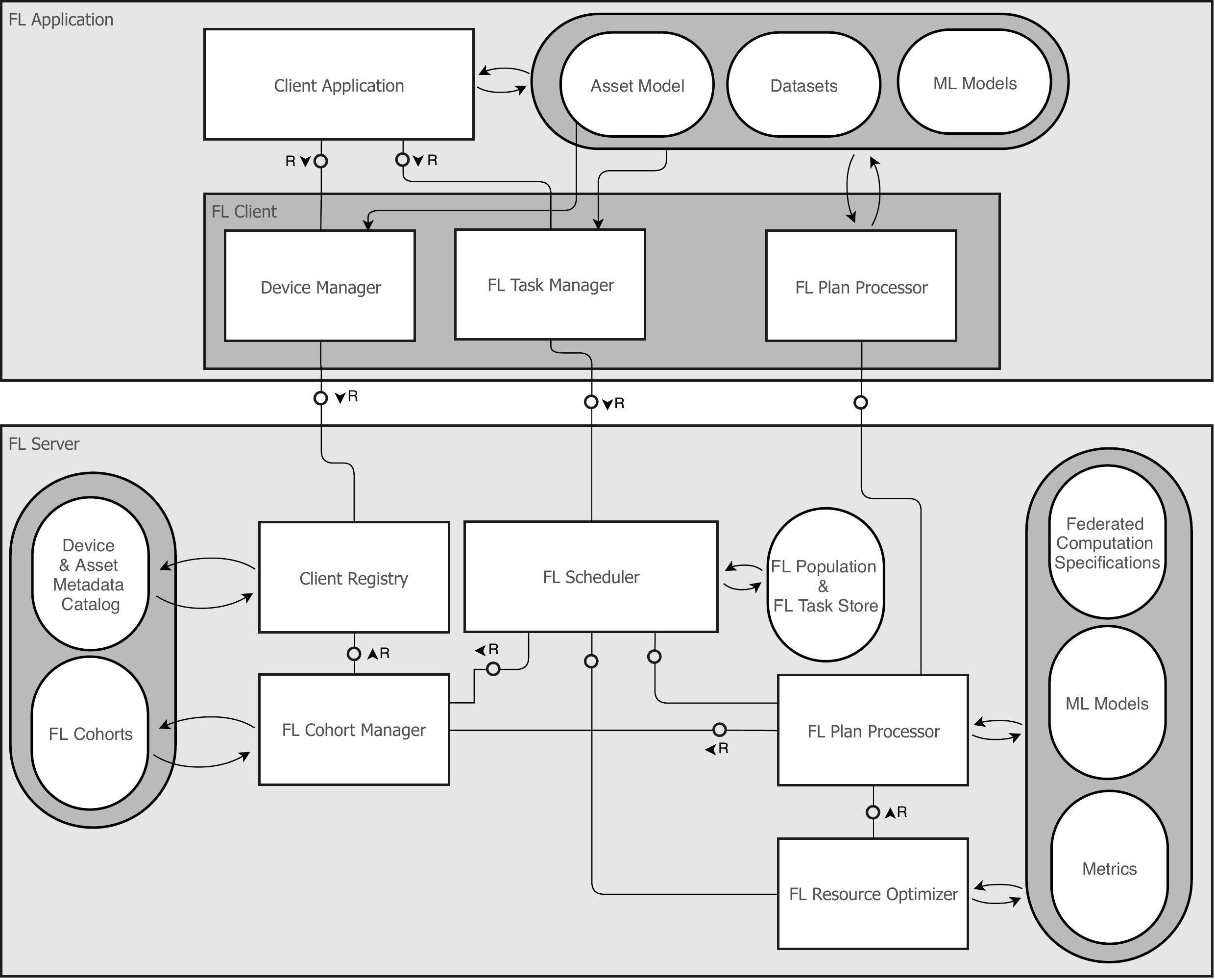}
    \caption{\acrshort{fl} Client and Server Architecture} \label{fig:Architecture}
\end{figure}

To realize the workflows presented in the previous section, we propose the \acrshort{ifl} architecture 
depicted in Figure~\ref{fig:Architecture}.

Considering two types of parties involved in \acrshort{ifl}, 
we present the \emph{FL Application} and the \emph{FL Server}, 
whereas the former is a container for a \emph{Client Application} that is a domain-dependent consumer of \acrshort{ifl}.
Furthermore, the \acrshort{fl} Application contains the \emph{FL Client} that interacts with the \acrshort{fl} server.

We now discuss the main components of the \acrshort{ifl} system and its responsibilities. 
First, the \acrshort{fl} client registration workflow involves the \emph{Device Manager} of the \acrshort{fl} client.
It provides an API to the client application to register for \acrshort{fl}. 
The client application provides a list of participating edge devices and general information of the organization.
Forwarding this to the \emph{Client Registry} 
allows persistence in the \emph{Device \& Asset Metadata Catalog} stored on the \acrshort{fl} server.
cohort search criteria posting is supported by device manager and client registry too,
with additionally exposing an interface to the \emph{FL Cohort Manager} to provide 
the device \& asset metadata catalog and the \acrshort{fl} cohort search criteria for creating \acrshort{fl} cohorts. 

Submitting new \acrshort{fl} tasks is initiated by invoking the \emph{FL Task Manager} 
which is in charge of enriching the information provided by the \acrshort{fl} task 
with information of the associated \acrshort{ml} model and targeted asset. 
After forwarding the \acrshort{fl} task to the server-side \emph{FL Scheduler}, 
it is mapped to the corresponding \acrshort{fl} population and persisted.
Furthermore, the FL Scheduler attaches scheduling information to timely trigger execution of all \acrshort{fl} tasks of a \acrshort{fl} population.
To actually run a \acrshort{fl} task, the \acrshort{fl} Scheduler hands it over to the \emph{FL Plan Processor}.
It translates the \acrshort{fl} task to a \acrshort{fl} plan and corresponding instructions 
as defined in \emph{Federated Computation Specifications}. 
Subsequently, it creates the corresponding global \emph{ML Model} and starts the \acrshort{fl} process for a given \acrshort{fl} cohort
by connecting to all \acrshort{fl} clients that have \acrshort{fl} tasks in the same \acrshort{fl} cohort. 
This information is provided by the \acrshort{fl} cohort manager. 
Analogously to \acrshort{fl} plans, there exists a client counterpart of the \acrshort{fl} plan processor too. 
It invokes the client instructions specified in the \acrshort{fl} plan 
to, e.g., train or evaluate \acrshort{ml} models on local edge devices.
\emph{Metrics} resulting from evaluation plans are provided by the \acrshort{fl} plan processor to the \acrshort{fl} cohort manager
to update cohorts continuously. Further metrics from, e.g., continuous learning approaches or \acrshort{qoi} evaluations 
are stored and used directly by the \acrshort{fl} plan processor e.g., during model aggregation.
The \emph{FL Resource Optimizer} connects to the metrics storage to incorporate parameters in optimization models.
After solving the optimization task, the solution is returned to the \acrshort{fl} scheduler to trigger, 
e.g., cohort reorganization and to update the schedule of \acrshort{fl} plan executions.

\section{Conclusions and Future Work}
\label{sec:conclusion}
In this work, we identified the need for \acrshort{ifl} and provided a structured collection 
of requirements and workflows covered in an \acrshort{ifl} architecture.
Due to diverse conditions of assets operating in industry, 
\acrshort{fl} clients are not advised to exchange \acrshort{ml} model parameters with the global set of \acrshort{fl} participants.
For this, we concluded to consider \acrshort{fl} tasks grouped in \acrshort{fl} cohorts 
aiming to share knowledge resulting from similar environmental and operating conditions of involved assets.  
Furthermore, we highlighed that \acrshort{fl} can decrease the amount of resource-intensive work of domain experts 
considering less continuous updates of datasets and labelling to be done. 
Additionally, making use of metrics resulting from \acrshort{qoi} and \acrshort{ml} model evaluations 
can be used for \acrshort{fl} cohort reorganizations and weighting in the \acrshort{fl} process.

As future work, we consider evaluation of a pilot implementation of the \acrshort{ifl} system in industrial labs.
Furthermore, the incorporation of \acrshort{fl} open source frameworks as \emph{PySyft}\footnote{https://github.com/OpenMined/PySyft},
\emph{TensorFlow Federated (TFF)}\footnote{https://www.tensorflow.org/federated}, and \emph{FATE}\footnote{https://fate.fedai.org/} 
needs to be evaluated with respect to production readiness and support for concurrent communication and computation needed for \acrshort{fl} cohorts.
Additionally, efficient asynchronous and decentralized \acrshort{fl} for industrial edge devices without involving a server is an interesting future research direction.
Finally, forecasting of potentially negative knowledge transfer that decreases model quality 
could complement the idea of dynamically reorganizing \acrshort{fl} cohorts.

\pagebreak
%
%
%
%

\bibliography{references}
\bibliographystyle{splncs04}

\end{document}